\documentclass[pdflatex,sn-mathphys-num]{sn-jnl}

\usepackage{tabularx}
\usepackage{array} 
\usepackage{graphicx}%
\usepackage{multirow}%
\usepackage{amsmath,amssymb,amsfonts}%
\usepackage{amsthm}%
\usepackage{mathrsfs}%
\usepackage[title]{appendix}%
\usepackage{xcolor}%
\usepackage{textcomp}%
\usepackage{manyfoot}%
\usepackage{booktabs}%
\usepackage{algorithm}%
\usepackage{algorithmicx}%
\usepackage{algpseudocode}%
\usepackage{listings}%

\usepackage{hhline}
\usepackage{tabularray}
\definecolor{ShipCove}{rgb}{0.427,0.556,0.776}
\definecolor{HalfBaked}{rgb}{0.58,0.784,0.788}
\definecolor{Tapestry}{rgb}{0.694,0.372,0.419}
\definecolor{WebOrange}{rgb}{1,0.647,0}
\definecolor{LightGray}{gray}{0.95}

\theoremstyle{thmstyleone}%
\theoremstyle{thmstyletwo}%

\theoremstyle{thmstylethree}%

\begin{document}

\title[Article Title]{From Norms to Indicators (N2I-RAG): An Agentic Retrieval-Augmented Generation Framework for Legal Indicator Computation}


\author*[1,2]{\fnm{Youssef} \sur{Al Mouatamid}}\email{youssef.almouatamid@univ-brest.fr}

\author[3]{\fnm{Marie} \sur{Bonnin}}\email{marie.bonnin@ird.fr}

\author[1,4]{\fnm{Jihad} \sur{Zahir}}\email{j.zahir@uca.ac.ma}

\affil[1]{\orgdiv{LISI Laboratory}, \orgname{Cadi Ayyad University}, \orgaddress{\city{Marrakesh}, \postcode{40000}, \country{Morocco}}}

\affil[2]{\orgdiv{LEMAR}, \orgname{Univ Brest}, \orgaddress{\city{Plouzane}, \postcode{F-29280}, \country{France}}}

\affil[3]{\orgname{IRD, Univ Brest, CNRS, Ifremer, LEMAR}, \orgaddress{\city{Plouzane}, \postcode{F-29280}, \country{France}}}

\affil[4]{\orgdiv{UMMISCO}, \orgname{IRD France Nord}, \orgaddress{\city{Bondy}, \postcode{F-93143}, \country{France}}}


\abstract{Computing legal indicators from normative texts is a key task in legal monitoring and policy evaluation, but presents significant challenges due to the complexity, scale, and interpretive nature of legal language, as well as the variability in available document quality. 
Existing natural language processing techniques and generative models can assist in legal analysis, but often suffer from high risk of hallucinations and lack the interpretability and evidence grounding required for reliable indicator computation. 
This paper presents N2I‑RAG (From Norms to Indicators), an agentic retrieval‑augmented generation framework designed to automate the computation of legal indicators in a transparent and traceable way. We integrate adaptive retrieval, llm-based agents, and validation mechanisms in a modular pipeline, where each component performs a defined role in filtering, retrieving, and assessing evidence, and in producing binary legal outcomes linked to identifiable legal provisions. The framework emphasizes traceability by requiring explicit explanations of intermediate decisions and final indicator assignments. We evaluate N2I‑RAG using an in-house constructed French marine environmental law corpus that includes both scanned and digital sources. Comparative experiments with multiple language model families demonstrate that the proposed approach consistently outperforms baseline systems, and generalizes well when tested on 2 different bans. The results indicate that agentic retrieval‑augmented generation can bridge open‑text legal language and standardized indicator computation, offering a foundation for transparent and scalable legal observatories.}




\maketitle

\section{Introduction}\label{sec1}

Legal indicators are structured measures that quantify legal implementation, compliance, or performance, serving as analytical tools for monitoring governance and policy outcomes.\\
In fields such as marine environmental law, legal indicators provide key evidence for assessing whether regulatory frameworks effectively address ecological challenges, from pollution to biodiversity loss. They translate legal obligations, often expressed in complex and qualitative terms, into standardized and comparable metrics usable by policymakers and researchers.\\
Several initiatives have demonstrated the value of legal indicators for environmental governance. For example, Michel Prieur et al. \cite{MeasuringtheEffectivityofEnvironmentalLaw} have developed an environmental law application indicator based on variables extracted from expert surveys. However, the construction of these indicators remains labor-intensive and depends on expert interpretation and extensive document analysis. As legal systems expand in scope and complexity, these practices face growing limitations in terms of scalability, consistency, and transparency. This creates a need for reliable and reproducible methods to compute legal indicators at scale.\\
The design of relevant legal indicators requires not only a clear conceptual framework but also a reproducible methodology and outcomes. Odeline Billant’s thesis \cite{billant2023} offers a comprehensive framework for legal indicator construction, emphasizing the alignment between legal norms, empirical observations, and operational metrics. This framework highlights the importance of traceability, as each indicator must be supported by explicit textual evidence.
However, implementing this framework in practice is challenging owing to the large -scale, heterogeneous, and linguistically complex legal texts. Manually identifying relevant norms and rules is both time-consuming and vulnerable to subjective interpretation.
Automating legal indicator construction through natural language processing techniques and generative AI is therefore a key element in enabling consistent, large-scale, and comprehensible monitoring of complex legal systems.\\
Natural Language Processing (NLP) has made progress toward automating aspects of legal analysis. Early work focused on tasks such as named entity recognition, document classification, and information retrieval have enabled partial automation to a certain degree of legal analysis workflows \cite{chalkidis-etal-2020-legal, oliveira2025, wang2025, costa2025, 10.3233/FAIA230972, 10.1007/978-3-319-99722-3_32, chen-etal-2020-joint-entity, pais-etal-2021-named}. While these methods support specific subtasks, they typically rely on limited annotated datasets and domain-specific rules, which limits their generalizability when faced with new topics or languages. Moreover, their systems often run as “black boxes”, offering little or no insight into the reasoning leading to how their outputs are produced. This lack of interpretability is particularly problematic for legal indicator computation, where transparency and justification are essential. In practice, models trained on court decisions or general legislation often underperform when applied to environmental law. Traditional NLP therefore remain insufficient for the level of consistency, interpretability and precision required in this context.\\
Large language models (LLMs) have taken natural language processing (NLP) to another level by enabling general purpose language understanding and generation, raising expectations regarding their potential contribution to legal analysis.\\
Their ability to perform zero-shot and few-shot reasoning makes them particularly appealing for legal text processing, where annotated data is scarce \cite{Breton2025}. Nevertheless, LLMs remain limited by their frozen training knowledge, their poor generalization to domain-specific queries, and, above all, their tendency to generate hallucinations, the phenomenon whereby the model outputs factually incorrect answers \cite{hal_surv, Huang_2025}. LLMs may confidently produce factually incorrect statements, lack up-to-date legal knowledge, and/or are insensitive to jurisdictional nuances.
Recent work \cite{RibeirodeFaria2025,youssef-llm} have demonstrated that, while LLMs can be valuable tools for information extraction, their outputs often lack factual grounding.
These studies show that even though LLMs have remarkable linguistic capabilities, their ungrounded nature limits their reliability for computing legal indicators, where every conclusion must be traceable and justified by verifiable evidence.\\
Retrieval-Augmented Generation (RAG) addresses some of these issues by grounding model outputs in external documents \cite{rag_ref}. Instead of encoding facts directly in model parameters, RAG systems retrieve relevant text snippets to support generation.
However, standard RAG pipelines follow a fixed sequence of retrieval and generation steps \cite{gao2024retrievalaugmentedgenerationlargelanguage}. This rigidity limits their ability to dynamically adapt queries, refine retrieval strategies, or evaluate the reliability of the intermediate results. Consequently, despite their potential, conventional RAG systems still lack the flexibility and adaptivity required to operate effectively in the legal field. Such approaches offer limited control over why and how legal texts are selected and evaluated which constrains their suitability for legal indicator construction.\\
Recent advances in agent-based systems extend RAG by introducing multiple reasoning steps, loops and workflows into these frameworks. These enhancements enable the agent to review and refine its own outputs.
In medicine, agentic RAG has improved factual reliability and interpretability \cite{zhao2025medragenhancingretrievalaugmentedgeneration}. However, in the legal domain, and particularly for legal indicator computation, these methods remains underexplored despite the need for explicit evidence tracking.
Extending agentic RAG to legal contexts represents a critical step toward building AI systems that can support trustworthy, interpretable, and adaptive legal analysis.\\
To address these limitations, this paper focuses on the automated construction of legal indicators from legal norms. 
We introduce N2I-RAG (From Norms to Indicators), an agentic retrieval-augmented framework designed to compute legal indicators from collections of digitized legal documents. It combines multi-agent coordination, semantic embeddings, and self-adaptive retrieval mechanisms to generate contextualized and traceable results. N2I-RAG formulates indicator computation as a structured information extraction problem, in which relevant texts are identified, then semantically aligned with Indicator queries, to be finally validated based on contextual relevance. These informations are then synthesized in a structured and traceable form, allowing experts to inspect how each indicator was derived. This approach aims to make legal indicator computation both automatable and interpretable, bridging qualitative legal analysis and quantitative monitoring.\\
The architecture is designed to be explainable and its outputs are interpretable in an operational sense. Each agent in the system produces, alongside its output, an explicit explanation that describes the reasoning process by which the result was obtained, including the sources used and the criteria applied. In this work, explainability denotes the availability of explanations at each stage of the pipeline, enabling inspection of intermediate decisions, while interpretability denotes the alignment of final outputs with the indicator definition and their direct linkage to identifiable legal articles. Requiring agents to produce such explanations constrains generation and supports consistent, evidence-based outputs.\\
The main contributions of this paper are:

\begin{enumerate}
    \item \textbf{Agentic RAG architecture:} A multi-agent pipeline that performs adaptive retrievals, contextual evaluations, query disambiguation, and evidence-based generation.
    \item \textbf{Evaluation and case study:} A comparative study and analysis of N2I-RAG’s performance against baseline RAG systems on marine environmental law.
    \item \textbf{Dataset construction:} A French marine environmental law corpus consisting of 10,596 legal articles drawn from five collections, covering legislation and regulations from both scanned and digital sources, structured and indexed to support retrieval-based analysis and indicator construction.
\end{enumerate}

Together, these contributions show how N2I-RAG can support the consistent and transparent computation of legal indicators across complex regulatory domains.

\section{Methodology}\label{sec2}

\begin{figure}[h!]
\centering
\includegraphics[width=0.99\textwidth]{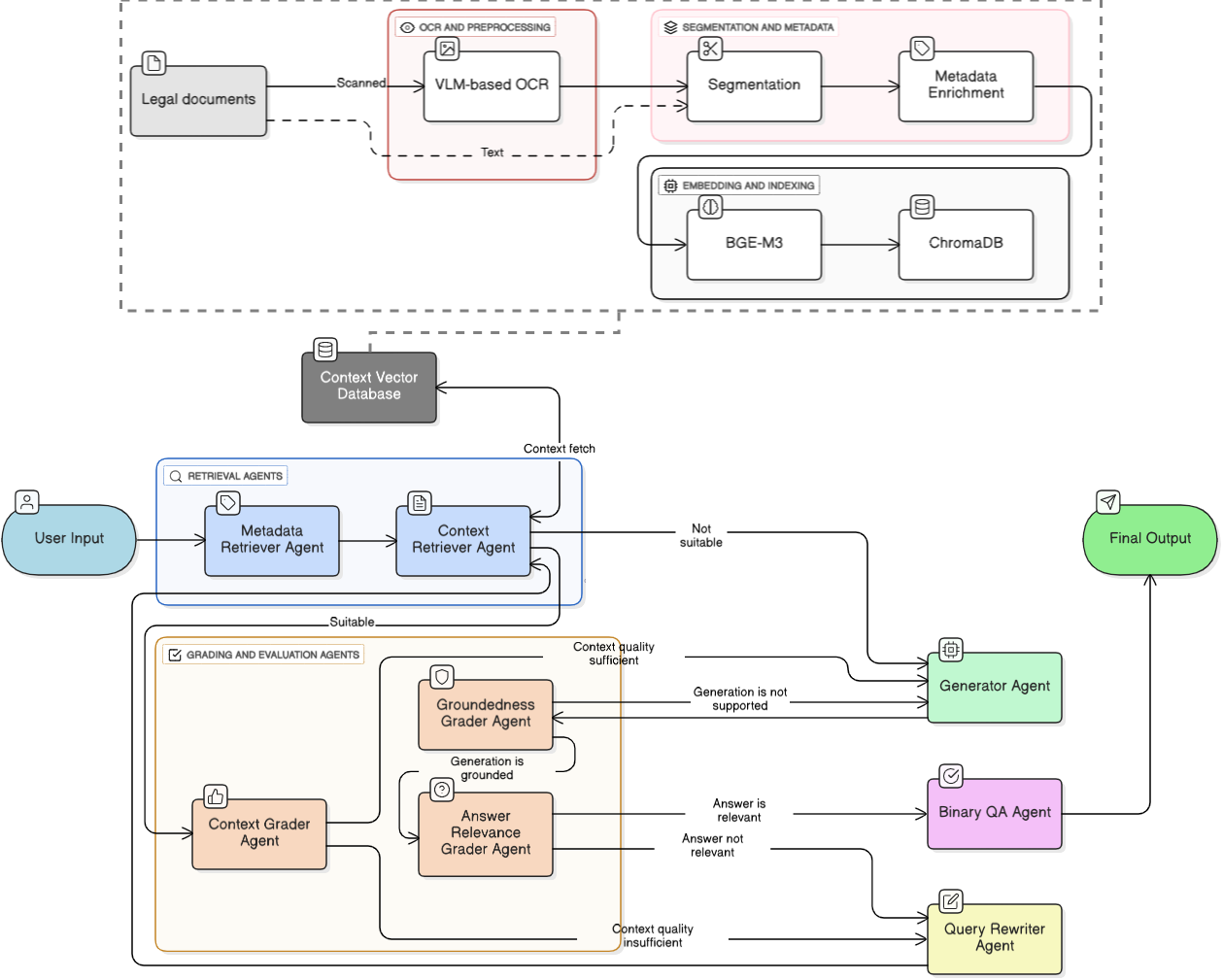}
\caption{Overview of the N2I-RAG Framework}\label{fig1}
\end{figure}

The N2I-RAG framework converts natural language legal queries into traceable binary decisions that populate an assessment grid representing legal indicators. The system connects document analysis to indicator calculation through a structured, multi-agent pipeline. Each step, from text extraction to decision-making, ensures that intermediate reasoning is visible and verifiable.
As illustrated in Figure \ref{fig1}, the process begins with OCR (Optical Character Recognition) processing of the document. The cleaned text is semantically indexed to enable accurate retrieval. An agentic reasoning loop then handles retrieval, grading, generating, and validating before returning a binary result. These outputs fill the evaluation grid, where each entry corresponds to a distinct legal indicator.
This design ensures that each computational step remains interpretable, traceable, and meets the legal reliability standards required for policy and compliance analysis.
\subsection{Visual-Language Processing}
Precise text extraction is vital for downstream legal analysis, especially when source documents include scanned and degraded documents.
Many legislative documents, notably in developing countries, are only available as low-quality scans with stamps, signatures, or inconsistent layouts. Conventional OCR often fails on these materials, leading to missing or corrupted text.
To address this, N2I-RAG uses a hybrid OCR strategy built on a Visual–Language Model (VLM) that integrates a vision encoder with a language decoder. This allows the system to capture both textual and visual elements (layout, formatting, and embedded elements) in order to recover full legal texts. The model corrects noisy segments, ignores irrelevant visual features, and generates missing text. Each document is then segmented into legal articles and enhanced with a set of standardized metadata such as jurisdiction, publication date, revision date, institutional source, and text type (law, regulation, directive,etc.). These articles are encoded using the BGE-M3 model [11] , a multi-lingual, multi-function, multi-granularity embedding model and indexed in ChromaDB for semantic search.
This approach guarantees that even imperfect documents provide usable data, producing a comprehensive, coherent, and searchable database suitable for indicator generation.

\subsection{Multi-Agent Framework}
The multi-agent pipeline structures N2I-RAG’s reasoning process into verifiable stages. Distributing tasks across specialized agents improves transparency and reduces cumulative errors. Each agent plays a specialized and complementary role, while recording its reasoning, so that errors and biases can be identified and addressed. This modular structure prevents information loss and reduces the risk of hallucination.
The pipeline includes eight different specialized agents. Table \ref{tab1} provides an overview
of these agents, their roles, and their interactions.
Collectively, these agents turn complex natural language reasoning into a transparent workflow featuring factual consistency and interpretability controls.
In the following subsections, we detail how each agent works individually, highlighting their specific contribution to controlling hallucinations and improving response quality.

\begin{table}[h]
    \centering
    \caption{Description of agents used in the architecture}\label{tab1}
    \begin{tabular}{>{\raggedright\arraybackslash}p{0.2\linewidth}>{\raggedright\arraybackslash}p{0.35\linewidth}>{\raggedright\arraybackslash}p{0.15\linewidth}>{\raggedright\arraybackslash}p{0.15\linewidth}}
    \toprule
      Agent & Role & Input & Output \\
    \midrule
     Metadata Retriever & Identify and parse metadata associated with a query & User query & Metadata JSON \\
     \midrule
     Context Retriever & Extract relevant articles & Query & Potential articles \\
     \midrule
     Context Grader & Evaluate the semantic relevance and specificity of the retrieved articles & Retrieved articles & Score + Explanation \\
     \midrule
     Generator Agent & Generate responses based on the selected context & Filtered and graded context & Natural language answer \\
     \midrule
     Groundedness Grader & Evaluates whether each claim is directly supported by cited context & Generated answer + Context & Score + Explanation \\
     \midrule
     Answer Relevance Grader & Check the consistency between the generated response and the initial request & Generated answer + query & Score + Explanation \\
     \midrule
     Query Disambiguator & Reformulate ambiguous or general queries & User query & Reformulated query \\
     \midrule
     Binary QA Agent & Assign a binary output (0/1) for the final evaluation grid & Generated answer & Binary label + Explanation \\
    \bottomrule
    \end{tabular}
\end{table}

\subsubsection{Metadata Retriever}
Metadata retrieval is an essential step in enriching the query and improving the accuracy of the results.
Metadata such as date, jurisdiction, or thematic provide contextual dimensions that complement purely textual information and enable more targeted searches.
In our framework, the module automatically identifies the metadata associated with a query and converts it to JSON, facilitating its use by subsequent agent. This standardized structure ensures consistency and interoperability within the system. Thus, the integration of metadata enhances the accuracy and contextualization of the retrieval process, paving the way for more appropriate responses.
\subsubsection{Context Retriever}
Cosine similarity is a particularly suitable approach for retrieving relevant documents in Natural Language Processing applications. When associated with ChromaDB and its HNSW (Hierarchical Navigable Small World) index, it offers a combination of high semantic accuracy and optimized performance at scale, which significantly improves the efficiency of the search system.\\
This measure evaluates the directional alignment of vectors in semantic space, regardless of their magnitude. This property neutralizes the text length dependency, which is essential when processing articles of mixed sizes and structures, a problem often encountered in information retrieval systems.\\
By using this metric with an HNSW index in ChromaDB, we benefit from approximate nearest neighbor (ANN) search, which makes the retrieval process fast and efficient, even for large data collections.\\
In our system, each article in our database is preprocessed and transformed into a vector representation using the BGE-M3 embedding model. These vectors are indexed in ChromaDB, which uses the HNSW index to perform cosine similarity searches. When a query is submitted, it is also converted into a vector and compared to document vectors using cosine similarity. The relevance score is calculated as shown by equation \ref{eq1}:

\begin{equation}\label{eq1}
d = 1.0 - \frac{\sum(A_i \times B_i)}{\sum(A_i^2) \times \sum(B_i^2)}
\end{equation}

Therefore, by using this equation \ref{eq1} as a distance function in ChromaDB's HNSW index, we optimize both semantic accuracy in article retrieval and the scalability and efficiency of the process, ensuring relevant and contextually appropriate item selection.
\subsubsection{Context Grader}
Context grading is crucial for sorting out the truly useful content from the superficially similar ones. Indeed, not all retrieved articles are equally relevant, and the linguistic complexity of legal texts can introduce noise, affecting the quality of the generated answers.\\
This agent grades each article based on two criteria: semantic relevance and specificity. Each grade is supported by a detailed explanation, ensuring transparency and interpretability.\\
By selecting and filtering the most informative contexts and justifying these decisions, this module guarantees the reliability and explainability of the data transmitted to the following steps, ensuring that their reasoning is based exclusively on reliable and contextually relevant information.
\subsubsection{Generator Agent}
The generator agent is the core of the process, as it produces the final response based on the selected information.
Controlled and contextualized generation is essential to avoid hallucinations and ensure that responses accurately reflect the content of the selected documents.
This module relies exclusively on filtered and evaluated texts. The generation process follows a strict set of rules to maintain consistency and factual accuracy while producing a coherent and understandable output. As a result, this module is able to provide accurate and reliable answers that are grounded in a validated context.
\subsubsection{Groundedness Grader}
We introduce a groundedness agent designed to check the factual reliability of responses and limiting hallucinations. A generative model can produce convincing statements without any source reference. However, without explicit grounding, the reliability of these assumptions is questionable, jeopardizing the credibility of the system. This module verifies that each statement in the final answer is explicitly linked to a text passage from the retrieved documents. Any statements without a source reference are flagged as unreliable. This approach allows a direct link to be traced between the source data and the generated content, enhancing the transparency of the system. Therefore, the grounding assessment acts as a safeguard against hallucinations, ensuring that each response element is based on documented evidence.
\subsubsection{Answer Relevance Grader}
Assessing the relevance of the response ensures that the output generated by the system remains consistent with the user's original intention. A response could appear factually correct and yet be irrelevant to the original query. Although this is not strictly a factual hallucination, it significantly reduces the usability of the system. This module systematically evaluates whether the initial query and the generated response are coherent. Hence, this module helps to ensure that the generated answers are not only correct, but also relevant and useful to the initial query.
\subsubsection{Query Disambiguator}
Query disambiguation and rewriting improves system performance by optimizing query formulation\cite{min-etal-2020-ambigqa}. Users may formulate queries that are ambiguous, incomplete, or too general. If not reformulated, retrieval is likely to be inefficient or produce too much noise.
This module automatically reformulates queries while preserving their initial intent while making them more specific and operational for retrieval. It thus reduces noise and improves the relevance of candidate articles. Query disambiguation is therefore an effective means of increasing the accuracy and coverage of the search system.
\subsubsection{Binary QA Agent}
The binary response module plays a strategic role in our framework, as it translates the responses generated into decisions suitable for integration into a 0/1 evaluation grid. In a setting where AI results are feeding a legal evaluation system, clarity and standardization of decisions are essential. Generative systems often produce nuanced or ambiguous responses that are difficult to encode in a binary format. However, to integrate them into a normative grid and ensure comparability between instances, we need to transform these nuances into a discrete signal (1 if the response is entirely affirmative, 0 otherwise). This step requires a level of decision-making rigor that goes beyond text generation.\\
The agent relies on two sets of instructions to guide its decision-making:
\begin{enumerate}
    \item Total affirmation vs. total negation: The response is evaluated as either entirely affirmative (1) or otherwise (0).
    \item Full coverage of the question: the agent assigns a score of 1 only if the answer covers all parts of the question, without omission or ambiguity. Any partial, contradictory, or incomplete answer automatically receives a score of 0.
\end{enumerate}

These guidelines are implemented via specific prompts, and the agent returns its verdict in JSON format containing both the binary score and a textual explanation.\\
The score is fed into the evaluation grid, while the explanation provides essential traceability for expert review and decision-making justification.\\
Consequently, the binary response module constitutes a bridge joining the flexibility of natural language and the normative strictness of the legal evaluation grid. By binding the system's outputs to a standardized binary format, it ensures the reliability, transparency, and applicability of its results as AI contributes directly to the legal evaluation.

The N2I-RAG framework combines multi-modal document processing and llm-based agents to translate unstructured legal texts into meaningful binary indicators. In each phase, from document processing to decision-making, N2I-RAG features a transparent and traceable workflow aligned with legal reliability standards.The integration of semantic retrieval, multi-agent and ground-based validation controls makes N2I-RAG stand out from other conventional solutions that prioritize performance over interpretability. This methodology paves the way for automated, evidence-based legal indicator computation, thus filling the existing gap between formal legal texts and quantifiable/measurable compliance outcomes.

\section{Experimental Setup and Case Study: Marine Environmental Law}
\subsection{Case Study Focus}
We evaluated N2I-RAG on a case study in marine environmental law, to assess its ability to automatically calculate reliable legal indicators from heterogeneous documents.
This domain is particularly well suited for evaluation due to the growing demand from international organizations for standardized legal indicators to monitor environmental protection. Marine environmental law is constantly expanding in both scope and complexity as governments and international institutions tackle urgent ecological crises, including climate change, biodiversity loss, and pollution. The process of evaluating these legal measures remains time-consuming and requires in-depth expertise, making it a appropriate testing ground for norm-to-indicator computation.
The study focuses on two prohibitions with measurable legal impact: 
\begin{itemize}
    \item The ban on plastic bags.
    \item The ban on bottom trawling.
\end{itemize}
Each ban is assessed through an evaluation grid for legal indicators covering 11 standardized questions, designed jointly by legal, ecological, and economic experts, detailed in the Table \ref{tab2}. These questions measure the applicability, characteristics, scope, and operationalization of the prohibition.
This setup therefore provides a clear and reproducible setting for testing N2I-RAG's performance in a real-world application, to translate legal norms into structured, indicator-level outputs.

\begin{table}[h]
    \centering
    \begin{talltblr}[
        caption = {The legal indicator is an agregation of 11 metrics. Blue and green metrics quantify the reach of the ban (scope extent and exception scarcity, respectively), while red and yellow represent operationalization (penalties and control mechanisms).},
        label = {tab2},
    ]{
      width = \linewidth,
      colspec = {X[l,m] Q[c,m]}, 
      rowsep = 6pt,
      row{1} = {font=\bfseries, fg=black, bg=LightGray},
      row{2-5} = {bg=ShipCove!20, fg=black},
      row{6-7} = {bg=HalfBaked!20, fg=black},
      row{8-9} = {bg=Tapestry!20, fg=black},
      row{10-12} = {bg=WebOrange!20, fg=black},
      hlines = {2pt, white}, 
      vlines = {white},
      cell{2}{2} = {r=11}{c, m, bg=white},
    }
    Evaluation metrics                                       & Output \\
    1. The ban is specified by a legal article                  & 0 or 1 \\
    2. The ban is nationwide in scope                           &        \\
    3. The ban is permanent in nature                           &        \\
    4. Details of banned activities is documented               &        \\
    5. There are no exceptions to the rule                      &        \\
    6. Exemptions are restricted to a few specific cases        &        \\
    7. A monetary fine is imposed                               &        \\
    8. A jail sentence is outlined                              &        \\
    9. A designated authority is tasked with enforcement        &        \\
    10. A control process with a defined duration is established &        \\
    11. A location-specific control procedure is detailed        &        
    \end{talltblr}
\end{table}

\subsection{Dataset Construction}
A consistent and high-quality corpus is essential for testing the reliability of retrieval-augmented legal analysis.
Legal documents are typically available as scanned images, with mixed quality and layout. Accurate computation of legal indicators requires both clean textual content as well as meaningful metadata. Our scope covers African countries where French is an administrative language.
Documents were acquired from FAOLEX \footnote{https://www.fao.org/faolex}, a database maintained by the Food and Agriculture Organization of the United Nations. It contains a vast collection of legislative, regulatory, and policy texts covering areas such as agriculture, natural resources, and the environment on a global scale. This source was chosen for its comprehensive coverage. This study focuses on African countries where French is used as an administrative language. 
We processed the corpus through the N2I-RAG preprocessing pipeline described in Section 2.2. All documents were processed and cleaned using the hybrid OCR system and the Visual–Language Model (VLM), which corrected any noise present in the text.
They were then segmented into articles, enhanced with the standardized metadata, and embedded using BGE-M3 representations.  Vectors were stored in ChromaDB for semantic retrieval.
Through this process, we have built a comprehensive database consisting of five distinct collections, totaling 10,596 articles, structured and indexed for research, analysis, and automated legal reasoning tasks, as shown in the Table \ref{tab3}.  We intentionally introduced three additional prohibitions to obtain a total of five collections in addition to the initial two, in order to challenge the retrieval process. Therefore, our vector database does not only contain articles related to our case study.

\begin{table}[h]
\centering
\caption{Number of Articles in the Corpus by Country and Ban}
\label{tab3}
\begin{tabular}{>{\raggedright\arraybackslash}p{0.12\linewidth}>{\centering\arraybackslash}p{0.16\linewidth}>{\centering\arraybackslash}p{0.12\linewidth}>{\centering\arraybackslash}p{0.14\linewidth}>{\centering\arraybackslash}p{0.14\linewidth}>{\centering\arraybackslash}p{0.1\linewidth}}
\toprule
\multirow{3}*{\textbf{Country}} & \multicolumn{5}{c}{\textbf{Ban}} \\
\cmidrule(lr){2-6}
                 & \textbf{Coastal Construction} & \textbf{Whaling} & \textbf{Bottom Trawling} & \textbf{Hydrocarbon discharge} & \textbf{Plastic Bags} \\
\midrule
Algeria       & 43   & 107  & 67   & 1,052 & -   \\
Benin         & 109  & 223  & 92   & 762   & -   \\
Cameroon      & -    & 105  & 105  & 93    & 24  \\
Congo         & -    & 155  & 16   & 246   & 11  \\
Ivory Coast   & 66   & 121  & 362  & 136   & 14  \\
Djibouti      & -    & 12   & -    & -     & -   \\
France        & -    & 13   & -    & -     & -   \\
Gabon         & 163  & 18   & 63   & 163   & 25  \\
Guinea        & 1,129& 469  & 33   & 2,538 & -   \\
Morocco       & 56   & 5    & 24   & 59    & 15  \\
Mauritania    & 701  & 101  & 67   & -     & 8   \\
Niger         & -    & -    & -    & -     & 15  \\
DRC           & -    & 91   & -    & -     & 17  \\
Senegal       & 363  & 138  & -    & -     & 42  \\
Togo          & -    & 28   & -    & -     & 18  \\
Tunisia       & 100  & 60   & -    & 147   & 6   \\
\midrule
\textbf{Total} & 2,730 & 1,646 & 829  & 5,196 & 195 \\
\bottomrule
\end{tabular}
\end{table}

This provides legally structured and semantically searchable references, offering a steady basis for indicator computation.

\subsection{Models and Implementation}
All experiments were executed locally to ensure reproducibility and control over resources. The implementation relies on LangChain and LangGraph for agent orchestration, and on Ollama for language model execution. All agents are implemented as specialized LLMs, with the exception of the retrieval module, which uses ChromaDB with bge-m3 vector representations and a search limited to the 10 closest results (top\_results=10).\\
We tested three state-of-the-art open-source large language models from different families: Llama3.2:3B, Qwen3:8b, and Mistral-Nemo:12b, in order to demonstrate the robustness of the proposed architecture.\\
Two generating settings were adopted:
\begin{itemize}
    \item For deterministic tasks (evaluation, scoring, binary decision), a temperature of 0 was used.
    \item For tasks requiring reformulation or creativity (rewriting queries, alternative generation), a temperature of 0.9 was used.
\end{itemize}

\subsection{Evaluation Metrics}
Performance was assessed according to standard binary classification metrics, but interpreted from a legal perspective:
\begin{itemize}
    \item \textbf{Accuracy:} Overall rate of correct answers (whether or not a law is correctly identified).
    \item \textbf{Sensitivity (Recall):} The model's ability to correctly identify situations where a law exists. A low recall would mean that effective laws are being ignored, which undermines the reliability of the legal diagnosis.
    \item \textbf{Specificity:} The system's ability to correctly identify cases where no law exists. This limits false positives, where a non-existent law is identified, which could be misleading
    \item \textbf{Precision:} Rate of responses correctly asserting the existence of a law among all those indicating that a law exists.
    \item \textbf{F1-Score:} Balance between precision and recall, reflecting the robustness of the system in the face of heterogeneous legal contexts.
    \item \textbf{Balanced Accuracy:} Average between sensitivity and specificity, helpful in cases of unbalanced data distribution (existing vs. non-existing law)
    \item \textbf{False Positive Rate (FPR):} Rate of falsely identifying a rule as existing.
    \item \textbf{False Negative Rate (FNR):} Rate of falsely identifying an existing rule as not existing.
\end{itemize}
This interpretation allows technical metrics to be grounded in a legal and operational interpretation, which is essential for regulatory assessment.

\subsection{Experimental Protocol}
The experiments were designed to evaluate both the global performance of N2I-RAG and the contribution of its key elements. By comparing full and reduced configurations, we can detect the impact of the different pipeline modules, especially those that reduce hallucinations.\\
Three configurations were tested:
\begin{enumerate}
    \item \textbf{Full Pipeline:} all agents are activated.
    \item \textbf{Retrieval-only Baseline:} Only vector search via ChromaDB is used, without LLM agents.
    \item \textbf{Pipeline without hallucination control agents:} Keep the main agents, except those dedicated to check relevance, groundedness, and coherence.
\end{enumerate}
Each configuration was run across the same indicators and both case study topics, ensuring a consistent comparison.\\
This ablation approach allows quantitative comparison and qualitative inspection, highlighting how agentic control improves factual grounding and stability, supporting a clear assessment of N2I-RAG’s overall reliability in legal contexts.

\section{Results and Discussion}
\subsection{Global Performance of N2I-RAG}
\begin{figure}[h!]
\centering
\includegraphics[width=0.9\textwidth]{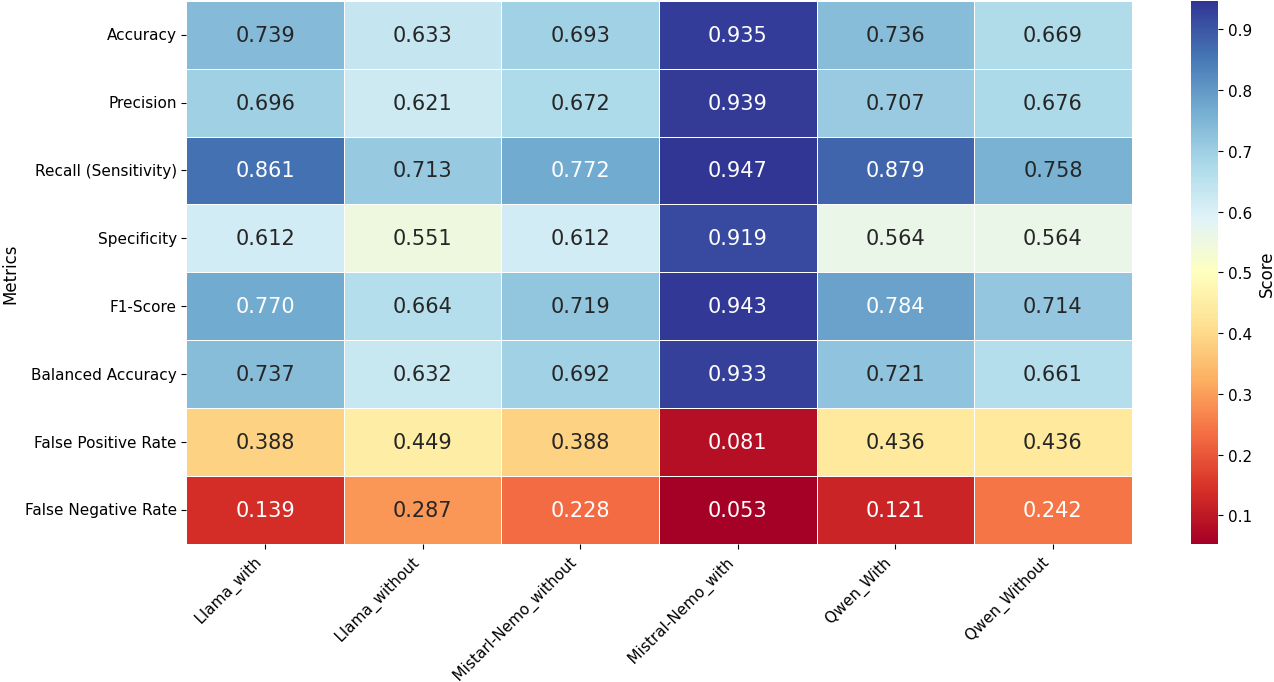}
\caption{Comprehensive and comparative metrics heatmap}\label{fig2}
\end{figure}

The experimental results (Figure \ref{fig2}) show how the integration of the N2I-RAG architecture significantly improves the quality of results across all LLM variants, with Mistral-Nemo:12B standing out in particular from the Qwen3:8B and Llama3.2:3B models.
The multi-agent pipeline reduces hallucinations, reinforces factual grounding, and ensures better consistency between retrieved legal texts and generated responses.
As shown in Figure \ref{fig2}, Mistral Nemo with N2I-RAG achieves a Balanced Accuracy of 0.933 and an F1-Score of 0.943, compared to only 0.768 without the architecture. Llama and Qwen also improve with N2I-RAG (0.77 and 0.784 in F1-Score, respectively), but remain behind. Specificity particularly illustrates this contribution: Mistral Nemo goes from 0.612 to 0.919.\\
These results confirm that the performance gains of N2I-RAG do not stem from the individual capacity of a model, but rather from its structured orchestration. The framework remains effective across architectures, proving that the proposed approach is LLM-agnostic but that its potential is maximized with more complex models, where combining representation ability and multi-agent control produces optimal output.
\subsection{Comparative Analysis Across Models}
N2I-RAG maintains stable performance across distinct LLM families, confirming its portability and robustness.
Because the proposed framework separates reasoning control from model output generation, it can adapt to models of different sizes and linguistic capacities. The architecture does not rely on memorized facts, but on active retrieval and validation loops.
While all models benefit, Mistral-Nemo consistently delivers the best results, suggesting that models with larger contextual windows are better at leveraging multi-agent interactions. Qwen and Llama still outperform their baselines, confirming that N2I-RAG’s gains are not model-dependent but rather result from structured orchestration.
This cross-model consistency supports N2I-RAG’s generalizability. It can be integrated into various LLM setups, enabling scalable and comprehensible automation for legal observatories.
\subsection{Ablation Study}
To assess the contribution of the core component, we conducted an ablation study comparing the full pipeline, a variant without hallucination control, and a simple baseline.
This analysis clarifies how each design component supports factual consistency and balanced performance

\begin{figure}[h!]
\centering
\includegraphics[width=0.9\textwidth]{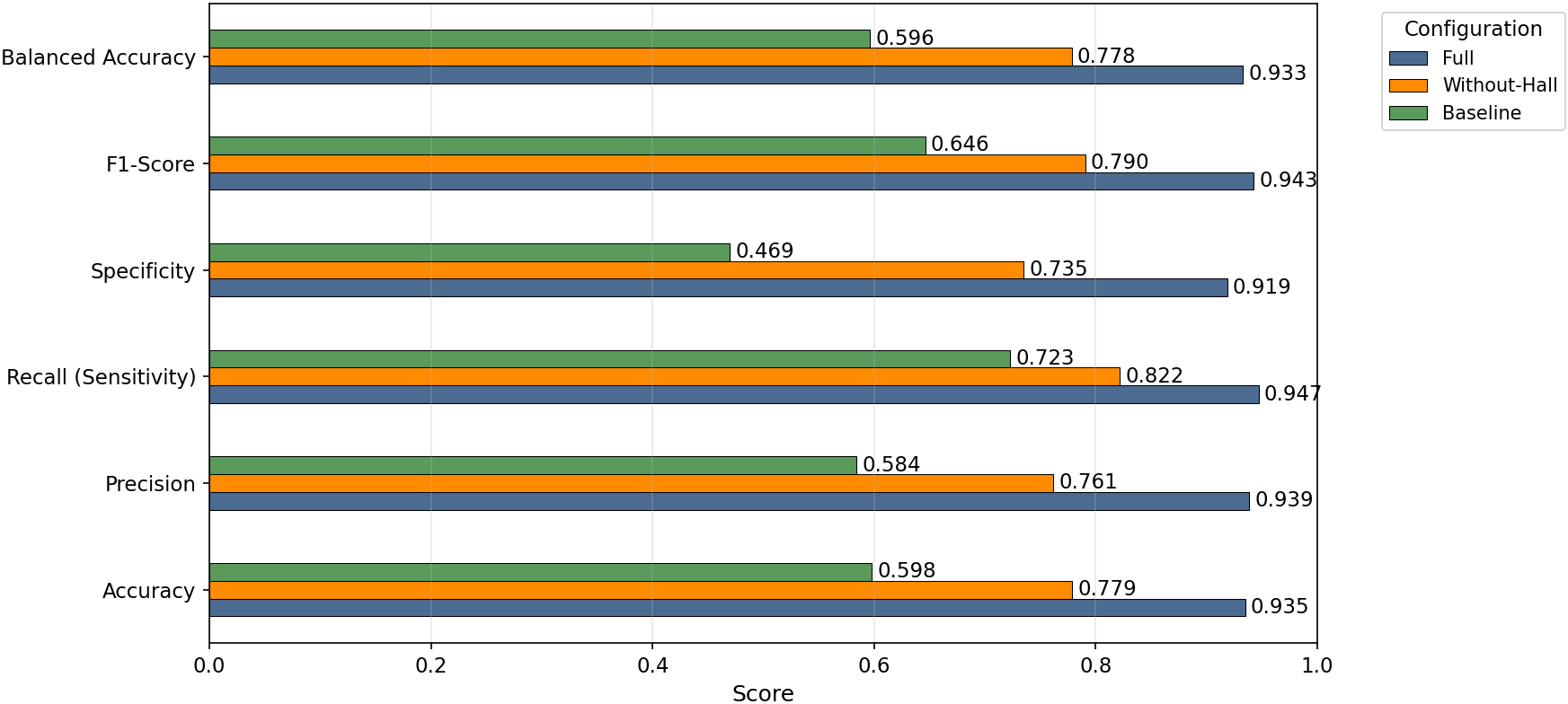}
\caption{Ablation test results for Mistral-Nemo by configuration}\label{fig3}
\end{figure}

\begin{table}[h]
\centering
\caption{Ablation Tests (Mistral-Nemo)}
\label{tab4}
\begin{tabular}{>{\raggedright\arraybackslash}p{0.07\linewidth}>{\raggedright\arraybackslash}p{0.14\linewidth}>{\raggedleft\arraybackslash}p{0.085\linewidth}>{\raggedleft\arraybackslash}p{0.085\linewidth}>{\raggedleft\arraybackslash}p{0.085\linewidth}>{\raggedleft\arraybackslash}p{0.085\linewidth}>{\raggedleft\arraybackslash}p{0.085\linewidth}>{\raggedleft\arraybackslash}p{0.085\linewidth}}
\toprule
\textbf{Ban} & \textbf{Configuration} & \textbf{Accuracy} & \textbf{Precision} & \textbf{Recall} & \textbf{Specificity} & \textbf{F1-Score} & \textbf{Balanced Acc.} \\
\midrule
\multirow{3}{*}{\begin{tabular}[c]{@{}l@{}}Plastic\\ bag\end{tabular}} & Full & 0.975 & 1.000 & 0.962 & 1.000 & 0.980 & 0.981 \\
 & Without-Hall & 0.843 & 0.862 & 0.848 & 0.836 & 0.855 & 0.842 \\
 & Baseline & 0.777 & 0.760 & 0.864 & 0.673 & 0.809 & 0.768 \\
\midrule
\multirow{2}{*}{\begin{tabular}[c]{@{}l@{}}Bottom\\ trawling\end{tabular}} & Full & 0.872 & 0.821 & 0.914 & 0.837 & 0.865 & 0.876 \\
 & Without-Hall & 0.782 & 0.765 & 0.743 & 0.814 & 0.753 & 0.778 \\
 & Baseline & 0.564 & 0.512 & 0.600 & 0.535 & 0.553 & 0.567 \\
\bottomrule
\end{tabular}
\end{table}
Figure \ref{fig3} and Table \ref{tab4} present the comparative results across three configurations obtained with the Mistral-Nemo model, selected for its superior performance. We note that the balanced accuracy for the plastic bag ban increased from 0.768 (Baseline) to 0.842 (Without-Hall) and then to 0.981 (Full). Similarly, for bottom trawling ban, performance improved from 0.567 (Baseline) to 0.876 (Full). Specificity is the most affected metric: with N2I-RAG, Nemo achieves 1.000 (plastic) and 0.837 (trawling), compared to only 0.836 and 0.535 without hallucination control.\\
These findings demonstrate that performance does not depend exclusively on the quality of LLM representations, but also on the design of the evaluation and control pipeline. The proposed multi-agent approach provides a key regulatory layer, particularly in sensitive contexts where legal false positives must be minimized.
\subsection{Legal Error Analysis}
The legal-oriented metrics enhance the relevance of the assessment by identifying two types of errors with different regulatory implications.\\
For the ban on plastic bags, Mistral-Nemo with N2I-RAG achieves Recall = 0.962 and Specificity = 1.000, illustrating a remarkable balance between recognition and reliability. Meanwhile, variants without anti-hallucination agents see their specificity drop to 0.535 in the case of trawling, which is a worrying false positive rate for legal applications.\\
The proposed architecture therefore provides a legal robustness guarantee by reducing critical biases related to false positives and false negatives. This outcome is particularly decisive when it comes to analyzing legal prohibitions in environmental law, where an erroneous claim can have serious regulatory and political implications.\\
Technical performance alone does not capture the practical reliability of legal AI systems. N2I-RAG was therefore analyzed from a legal-error perspective, focusing on the types and implications of false predictions.
Despite its robustness, N2I-RAG’s remaining errors reveal important insights about the nature of legal data and the boundaries of automated interpretation. For instance, in assessing the ban on plastic bags, the errors observed are not solely due to technical limitations of language models, but also reflect specific challenges related to the interpretation of the standard corpus used.\\
Two main categories of legal errors have been identified:
\begin{enumerate}
    \item \textbf{Data gaps:} In some cases (e.g., Ivory Coast), the sanctions associated with the ban are not mentioned in the articles accessible via the database, but appear in a separate decree which is not listed in Faolex. The lack of this secondary document led the system to incorrectly conclude that there were no sanctions.\\
    From a technical standpoint, this is not an error but a positive outcome, as the system did not produce incorrect information. However, from a legal perspective, this highlights the challenges of evaluating a law based on partial sources. It also shows the need to consider supplementary texts (like decrees and implementing orders), while emphasizing their limited online accessibility.
    \item \textbf{Transitional misinterpretation:} The model interpreted the phase-out period for plastic bags in the Democratic Republic of Congo as a temporal constraint on the ban. However, from a legal perspective, this measure is transitional and aims to regulate the progressive implementation of the ban.
\end{enumerate}
In legal terms, false positives mean over-claiming a regulation that does not exist, while false negatives omit and ignore an actual ban. Both have judicial consequences: the former can mislead compliance assessments, while the latter can hide regulatory progress.\\
These outcomes confirm that automating legal indicator computation requires both reliable data sources and structured interpretability mechanisms. N2I-RAG effectively reproduces expert reasoning, but depends on the comprehensiveness of accessible legal texts. Its results thus provide a transparent, evidence-based foundation for legal observatories aiming to systematically monitor norms, extending the conceptual foundations proposed by Billant (2023)\cite{billant2023} to an AI-supported operational workflow.
\section{Conclusion}
This paper introduced N2I-RAG, an agentic retrieval-augmented generation framework designed to compute legal indicators in a reliable, interpretable, and automatable way. 
It shows that structured multi-agent reasoning can bridge the gap between open-textured legal language and standardized assessment grids used in policy and judicial evaluation.\\
N2I-RAG integrates search, generation, and validation within a modular pipeline. Each agent performs a specific function, whether it be filtering, retrieving, grading, or producing binary legal results. This structure allows the system to analyze complex legal documents while maintaining full traceability of its decisions and reasoning process. Unlike conventional RAG systems that rely on a single generation step, N2I-RAG applies multi-level control to align the model’s linguistic output with the logic of legal indicators.\\
Experimental results demonstrate that the complete N2I-RAG architecture substantially outperforms its simplified variants, especially in sensitivity and specificity, two metrics essential for legal interpretation. Mistral-Nemo 12B achieved the highest performance, while results from Qwen and LLaMA confirm that the framework remains model-agnostic. Additionally, error analysis reveals that the challenges encountered arise primarily from the subtleties of the law (e.g., transitional periods, gaps in documentation, or implicit exceptions) rather than technical shortcomings , thus paving the way for further interdisciplinary work.\\
Beyond its technical performance, N2I-RAG contributes to the broader goal of automating legal observatories. It provides a foundation for continuous, transparent, and scalable monitoring of environmental law. This would offer ongoing legal monitoring, enhance visibility, and provide a practical decision-making tool for researchers, legal professionals, and even the general public. For future work, we intend to extend this framework to multilingual and multi-jurisdictional contexts, integrating richer databases and real-time data sources to support more adaptive and inclusive legal indicator computation.

\bmhead{Acknowledgements}
This work has been supported by AIME and ISblue project, Interdisciplinary graduate school for the blue planet (ANR-17-EURE-0015) and co-funded by a grant from the French government under the program \textit{”Investissements d’Avenir”} embedded in France 2030.


\bibliography{my_bibliography}

\end{document}